\documentclass{AISB2008}

\usepackage{balance}  
\usepackage{graphics} 
\usepackage{times}    
\usepackage{url}      
\usepackage{cite}
\usepackage{amsmath,amssymb,amsfonts}
\usepackage{algorithm}
\usepackage{algpseudocode}
\usepackage{mathtools}
\usepackage{bm}
\usepackage{caption}
\usepackage{subcaption}

\makeatletter
\def\url@leostyle{%
  \@ifundefined{selectfont}{\def\UrlFont{\sf}}{\def\UrlFont{\small\bf\ttfamily}}}
\makeatother
\newcommand{\beginsupplement}{%
	\setcounter{table}{0}
	\renewcommand{\thetable}{S\arabic{table}}%
	\setcounter{figure}{0}
	\renewcommand{\thefigure}{S\arabic{figure}}%
}
\def\pprw{8.5in}
\def\pprh{11in}

\usepackage[pdftex]{hyperref}
\hypersetup{
pdftitle={SIGCHI Conference Proceedings Format},
pdfauthor={LaTeX},
pdfkeywords={SIGCHI, proceedings, archival format},
bookmarksnumbered,
pdfstartview={FitH},
colorlinks,
citecolor=black,
filecolor=black,
linkcolor=black,
urlcolor=black,
breaklinks=true,
}

\makeatletter
\def\blfootnote{\xdef\@thefnmark{}\@footnotetext}
\makeatother

\begin{document}

\title{Weighting NTBEA for Game AI Optimisation}

\author{James Goodman \and Simon Lucas \institute{Queen Mary University of London, Game AI Reseach group, email: james.goodman@qmul.ac.uk} }

\maketitle

\begin{abstract}
The N-Tuple Bandit Evolutionary Algorithm (NTBEA) has proven very effective in optimising algorithm parameters in Game AI. A potential weakness is the use of a simple average of all component Tuples in the model. This study investigates a refinement to the N-Tuple model used in NTBEA by weighting these component Tuples by their level of information and specificity of match. We introduce weighting functions to the model to obtain Weighted-NTBEA and test this on four benchmark functions and two game environments.
These tests show that vanilla NTBEA is the most reliable and performant of the algorithms tested. Furthermore we show that given an iteration budget it is better to execute several independent NTBEA runs, and use part of the budget to find the best recommendation from these runs.
\blfootnote{\\ Accepted as a paper at the 11th AI and Games Symposium at AISB 2020}
\end{abstract}

\section{Introduction}
In Game AI, as in many other fields, algorithms usually have several parameters that need to be specified. For any given problem some parameter settings may give good results, while other settings give very poor results. For any new problem (a new game for example) we need to decide on which parameter values to use. In many cases a set of `standard' parameter settings are available based on previous work, but these may not be ideal for the new domain. An exhaustive search of all possible parameter settings is usually unfeasible - it may take days of processing time on a large parallel cluster to train a complex neural network using Reinforcement Learning (RL). If the RL algorithm has four parameters, each of which can have five values, then training a policy under each possible setting will take $5^{4}$ = 625 cluster-days, or about 2 cluster-years to evaluate each. 
The problem is considerably worse if the outcome of any one evaluation (or experiment) is stochastic, so that a good estimate of the value of a given parameter setting requires many independent evaluations. 

The field of parameter (and hyper-parameter) optimisation seeks fast methods for deciding on parameter settings in a new domain with an available computational budget. This generally involves constructing a predictive model for the result of a future untried evaluation. After each time-consuming real-world evaluation has been run, the computationally cheap predictive model is updated with the result and interrogated to suggest the next set of parameter values to try. By reducing the number of expensive full evaluations to find a good (if not necessarily optimal) set of parameters, we save significant time and money.

The N-Tuple Bandit Evolutionary Algorithm (NTBEA) was introduced in \cite{Kunanusont_Gaina_Liu_Perez-Liebana_Lucas_2017, Lucas_Liu_Perez-Liebana_2018}. It has been benchmarked against several other optimisation algorithms in stochastic game environments and proven to be more effective at finding a good set of parameter settings than other algorithms within a fixed computational budget \cite{Lucas_Liu_Bravi_Gaina_Woodward_Volz_Perez-Liebana_2019}. Similarly \cite{Sironi_Winands_2019} find NTBEA is the best optimiser of a number tried modify MCTS parameters during algorithm execution for a number of games.

NTBEA in \cite{Lucas_Liu_Perez-Liebana_2018, Lucas_Liu_Bravi_Gaina_Woodward_Volz_Perez-Liebana_2019} estimates the value of a set of parameter values using the simple average of all matching Tuples in the model (see Background for a detailed explanation). The current work extends this to weight the matching Tuples using the amount of data (i.e number of real-world experiments) that inform a given Tuple, and the degree of specificity of the Tuple match. We hypothesise that this approach will allow us to converge to a good parameter setting faster and more robustly than vanilla NTBEA. 

In addition to introducing Weighted-NTBEA in this work, we also modify four benchmark tests from the function optimisation literature to incorporate noise. These enable optimisation algorithms to be compared cheaply (in terms of computational budget) and also provide greater confidence in conclusions because the true underlying value is known exactly, and is not an estimate over multiple expensive evaluations. 

\section{Background} \label{Background}
\subsection{Black-box optimisation}
Black-box function optimisation addresses the problem of finding the optimal value of some $f(\theta)$
\begin{equation}
y = \max_\theta{ f(\theta)} \ \ \ \ \theta \in \mathbb{R}^d
\end{equation}
where $f(\theta)$ can be evaluated at any $\theta$, but not differentiated. When $f(\theta)$ is expensive to evaluate we wish to minimise the number of evaluations we make and can use the real evaluations made so far to model the result of $f(\theta)$ (the `response surface') to decide what value of $x$ should be evaluated next. A common approach is to use Bayesian optimisation techniques with a prior over the response surface, and update a posterior model after each evaluation. To pick the next point a trade-off is made between exploitation and exploration; for example the point with the largest expected improvement (EI), or the highest 95\% confidence bound (UCB) \cite{Shahriari_Swersky_Wang_Adams_deFreitas_2016, Brochu_Cora_deFreitas_2010, Jones_Schonlau_Welch_1998}. Bayesian methods require either a model to be specified, or a decision on the kernel functions to use in a (non-parametric) Gaussian Process. They are sensitive to stochastic noise, especially noise that is highly non-Gaussian \cite{Brochu_Cora_deFreitas_2010}. Approaches exist to integrate different types of noise into the model, but these add complexity to the model \cite{Shahriari_Swersky_Wang_Adams_deFreitas_2016}. 

Most Bayesian methods and libraries assume that $\theta$ is continuous in all dimensions $d$, and do not work in discrete spaces. This is not true for all, for example BOCS \cite{Baptista_Poloczek_2018} uses Bayesian Linear Regression with semi-definite programming to optimise a discrete combinatorial problem. However, BOCS does assume uniform Gaussian noise.
Other approaches have been used to model the response surface in black-box optimisation: Random Forests are used in the SMAC algorithm \cite{Hutter_Hoos_Leyton-Brown_2011}.

In a bandit-based approach, each setting of the parameters is one `arm' of the bandit, and we seek to find out which `arm' gives us the highest reward in a limited number of pulls. 
This is a natural fit if each $\theta_i$ can take a small number of discrete values, but it cannot cope easily with continuous dimensions. 

NTBEA combines a bandit-based approach with an N-Tuple model \cite{Kunanusont_Gaina_Liu_Perez-Liebana_Lucas_2017} and an evolutionary algorithm to select the next point to be evaluated. The UCB1 (Upper Confidence Bound) algorithm  is used to balance exploration and exploitation  \cite{Auer_Cesa-Bianchi_Fischer_2002}. 
NTBEA is described in detail in the next section.


\subsection{NTBEA} \label{NTBEA}
This explanation of NTBEA closely follows \cite{Lucas_Liu_Perez-Liebana_2018}.
During each iteration of NTBEA we:
\begin{enumerate}
	\item{Run a full game (or experiment, or other expensive function evaluation) using the current test setting $\bm{\theta}$. For the first iteration $\bm{\theta}$ is selected at random.}
	\item{Update the N-Tuple Model with the evaluation result.}
	\item{Generate a neighbourhood of points by applying a mutation operator to $\bm{\theta}$ (repeat $X$ times to get a neighbourhood of size $X$). }
	\item{Evaluate the Upper Confidence Bound (UCB) for each of the N points using the N-Tuple Model. Select the one with the highest UCB as the new $\bm{\theta}$, and repeat from 1.}
\end{enumerate}
In this study, as in  \cite{Lucas_Liu_Bravi_Gaina_Woodward_Volz_Perez-Liebana_2019, Lucas_Liu_Perez-Liebana_2018, Kunanusont_Gaina_Liu_Perez-Liebana_Lucas_2017} we set $X$=50, and the mutation operator used is to randomly mutate each $\theta_i$ to a random setting with probability $\frac{1}{d}$, always mutating at least one $\theta_i$.
\subsubsection{N-Tuple Model} \label{NTuple}
An 1-Tuple model breaks down the modelled $f(\bm\theta)$ into $d$ components, where $\bm\theta \in \mathbb{R}^d$ using Equation(\ref{1Tuple}). Each component $i$ is the expected value of $f$ assuming that only $\theta_i$ affects the value. If $\theta_i=x$, this is the mean of all evaluation results so far where $\theta_i=x$. In (\ref{1Tuple}), $\bm{1}(\theta_i = \phi_i)$ is a delta-function that is 1 when a previously evaluated $\bm\phi$ matches with the current $\bm\theta$ setting in the $i$th dimension, $N$ is the total number of previous evaluations, and $f_{obs}(\bm\phi_m)$ is the $m$th of these. $M_i$ is the number of evaluations that match with tuple $i$.
\begin{equation} \label{1Tuple}
\hat{f}(\bm\theta)= \frac{1}{d}\sum_{i=1}^d{\frac{1}{M_i} \sum_{m=1}^{N}{ \bm{1}(\theta_i = \phi_i)f_{obs}(\bm\phi_m)}}
\end{equation}
\begin{equation} \label{M_i}
\text{where}\; M_i =  \sum_{m=1}^{N} \bm{1}(\theta_i = \phi_i)
\end{equation}
In other words, our prediction $\hat{f}(\bm\theta)$ is the average of all the $d$ matching 1-Tuple predictions based on past observations. There are no interactions between different parameters, and there are no assumptions about relationships between different values of a given parameter. For example, if one parameter has discrete values 1, 2 or 3 then the result of evaluations where this was 1 or 3 will have no impact at all on predictions for the intermediate 2. This is a very conservative non-parametric model. 
In the case of 5 dimensions with 10 possible values for each, we need to maintain just 50 sets of statistics for a 1-Tuple model ($M$, the number of times each tuple-setting has been tried, and $\bar{f}$, the mean of these evaluations). Any $\bm\theta$ will match with exactly five of these, and $\hat{f}(\bm\theta)$ is the mean of these five.

A 2-Tuple model extends this to consider interactions between two parameter settings. We replace  $\bm{1}(\theta_i = \phi_i)$ in (\ref{1Tuple}) with $\bm{1}(\theta_i = \phi_i, \theta_j = \phi_j)$, and now consider all evaluations that were a match on two different parameters. In the case of 5 dimensions with 10 possible values for each this gives a total of $\binom{5}{2} \times 10 \times 10 = 1000$ distinct 2-Tuples for which $N$ and $\bar{f}$ are maintained.  Any $\bm\theta$ will match with exactly $\binom{5}{2}=10$.

In all the experiments in this study, as in \cite{Lucas_Liu_Bravi_Gaina_Woodward_Volz_Perez-Liebana_2019, Lucas_Liu_Perez-Liebana_2018, Kunanusont_Gaina_Liu_Perez-Liebana_Lucas_2017} we use 1-Tuples, 2-Tuples and $d$-Tuples in the model. A $d$-Tuple matches on all parameters, so is unique for each $\bm\theta$.
The predicted value $\hat{f}$ of the model for any new $\bm\theta$ is the arithmetic mean across \emph{all} matching tuples. 

\subsubsection{UCB} \label{UCB1}
The UCB1 algorithm \cite{Auer_Cesa-Bianchi_Fischer_2002} calculates a probable upper bound on the true value $J$ of the `arm' of a bandit $\bm\theta$, given the data observed so far using (\ref{UCB}). $N$ is the total number of trials of the bandit, and $n(\bm{\theta})$ is the number of times this `arm' has been pulled (i.e. the number of times that $\bm\theta$ has been evaluated).
\begin{equation}\label{UCB}
J(\bm{\theta}) = \hat{f}(\bm{\theta}) + k \sqrt{\frac{\log{N}}{n(\bm{\theta})}}
\end{equation}
The N-Tuple model uses equation (\ref{1Tuple}) to calculate $\hat{f}(\bm\theta)$, but we still have the second term of equation (\ref{UCB}) that controls exploration. 
We can calculate this for each individual tuple, with $N$ equal to the total number of NTBEA iterations, and $n(\bm\theta)$ equal to the number of these for which the tuple matches $\bm\theta$ in (\ref{1Tuple}).
NTBEA calculates the second term for each matching tuple, and then takes the arithmetic average. There is one additional nuance that some tuples will never have been evaluated, and formally (\ref{UCB}) will return $\infty$ in this case. To avoid this an additional hyper-parameter $\epsilon$ is added, so that 
\begin{equation}\label{UCBEpsilon}
J(\bm{\theta}) = \hat{f}(\bm{\theta}) + k \sqrt{\frac{\log{N}}{n(\bm{\theta}) + \epsilon}}
\end{equation}
In this study, as in  \cite{Lucas_Liu_Bravi_Gaina_Woodward_Volz_Perez-Liebana_2019, Lucas_Liu_Perez-Liebana_2018, Kunanusont_Gaina_Liu_Perez-Liebana_Lucas_2017} we set $\epsilon=0.5$. The value of $k$ needs to be scaled to the range of $f(\bm\theta)$, and is set for each domain (see Method section).

\section{Hypothesis} \label{Hypothesis}
Vanilla NTBEA estimates the value of a parameter setting $\bm{\theta}$ as the simple arithmetic mean of all the matching Tuples in the model that match. For example if we have five parameters and are using 1-, 2- and N-Tuples then any $\bm{\theta}$ will have one matching $d$-Tuple (where $d=5$), five matching 1-Tuples and $\binom{5}{2}=10$ matching 2-Tuples. The statistics gathered for each of these 16 Tuples is then averaged. The same approach applies to calculating the exploration estimate using (\ref{UCB}).
Even if we have evaluated a specific $\bm{\theta}$ multiple times, the results from those evaluations still only comprise $\frac{1}{16}$ of the NTBEA estimate; $\frac{5}{16}$ always comes from the matching 1-Tuples. Our hypothesis is that NTBEA will better estimate the value of a parameter setting if it applies greater weight to the more specific tuples as the number of evaluations increases. In the limit of a large number of evaluations of a specific $\bm{\theta}$, then only the statistics from the fully-matching $d$-Tuple should be relevant.

We propose four distinct weighting schemes, which vary in the rate of decay in the influence of less-specific tuples. In all cases the value $V$ of a parameter setting $\bm{\theta_N}$, with $N$ different parameters is
\begin{equation} \label{genericWeight}
V(\bm{\theta}_N) = w TP(\bm{\theta}_N) + (1-w)\frac{1}{|\bm{\theta}_{N-1}|}\sum{V(\bm{\theta}_{N-1}) }
\end{equation}
where $TP(\bm{\theta}_N)$ is the average value from the N-Tuple statistics of $\bm{\theta}_N$ and $w \in [0, 1]$ is the weight used for the N-Tuple statistics. The remaining 1-$w$ weight is applied to the average of all (N-1)-Tuples, i.e. all Tuples on the next level down. In (\ref{genericWeight}), $|\bm{\theta}_{N-1}|$ is a slight abuse of notation and refers to the number of such Tuples. In the case that no Tuples are held at the N-1 level, then this descends to the next level for which we do have Tuples in the NTBEA model. Note that (\ref{genericWeight}) is recursive, and each of the $V(\bm{\theta}_{N-1})$ terms is calculated from weighting its Tuple statistics, $TP(\bm{\theta}_{N-1})$, with a sum over $V(\bm{\theta}_{N-2})$ at the next level down.

In our $N=5$ example vanilla NTBEA always weights the 5-Tuple, the ten 2-Tuples and the five 1-Tuples at $\frac{1}{16}$ each. Using (\ref{genericWeight}) this weighting will change as we gain more information. With no evaluations, the $w$ for any Tuple will be $0.0$, and as the number of evaluations for a Tuple increases we want $w$ to increase towards a maximum of $1.0$ so that asymptotically we ignore information from lower-level Tuples.

The four weighting schemes use linear, inverse, inverse square-root and exponential decay functions.

\begin{figure}[!t] 
	\centering
	\includegraphics[width=3in]{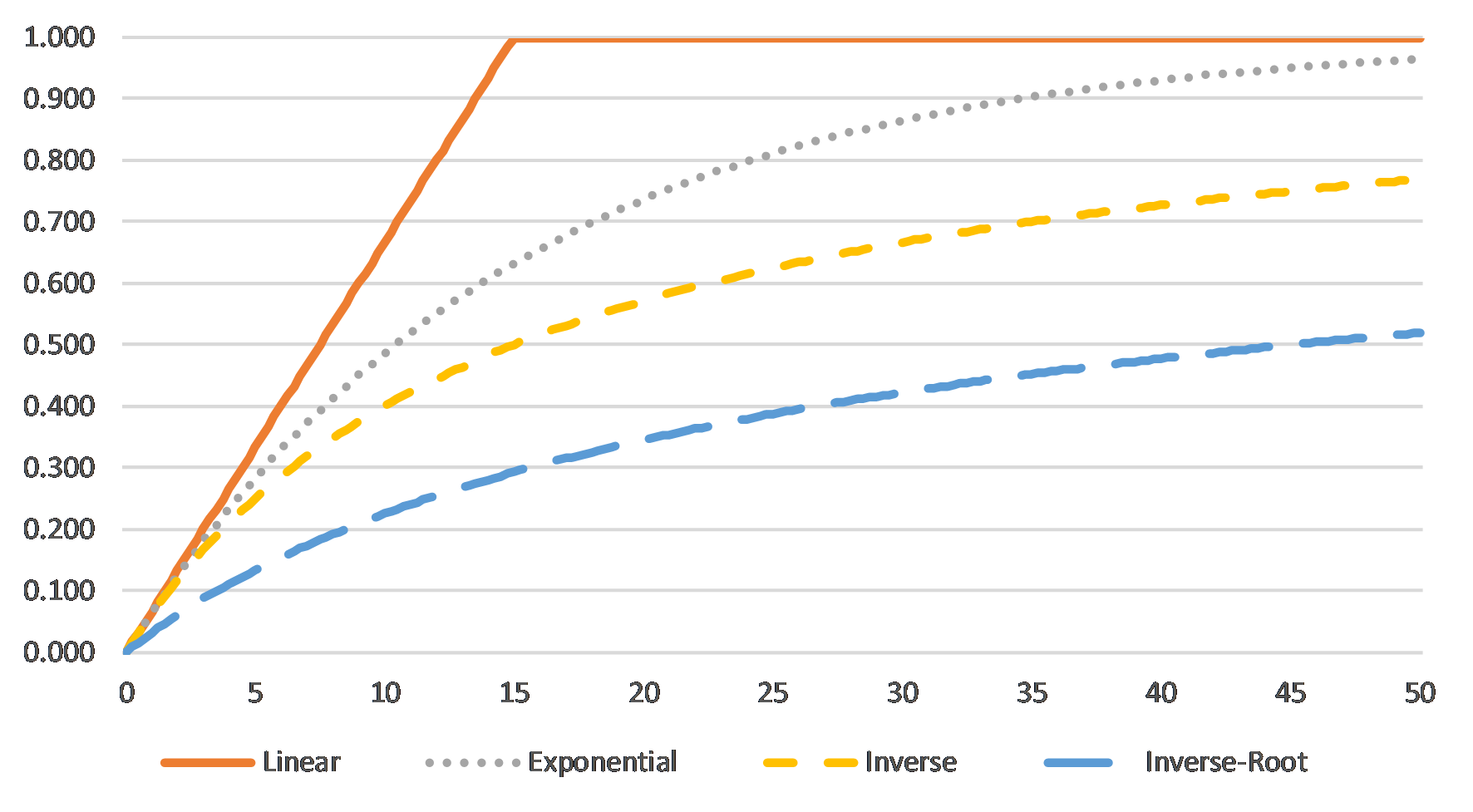}
	\caption{The four weighting functions used in Weighted-NTBEA. The x-axis is $n(\theta)$ the number of evaluations that match the tuple, and the y-axis is the weighting applied to the tuple between 0 and 1. The remainder of the value is calculated from the average of the next level of tuples. $T=15$ in all cases.}
	\label{functionPicture}
\end{figure}

\begin{enumerate}
	\item{Linear
	\begin{equation} \label{linear}
	w =  \min\left(\frac{n(\theta)}{T}, 1.0\right)
	\end{equation}}
	\item{Inverse-root
	\begin{equation} \label{inverseRoot}
	w = 1.0 - \sqrt{\frac{T}{n(\theta) + T}}
	\end{equation}}
	\item{Inverse
	\begin{equation} \label{inverse}
	w = 1.0 - \frac{T}{n(\theta) + T}
	\end{equation}}
	\item{Exponential
	\begin{equation} \label{exponential}
	w =  1.0 - \exp\left(\frac{-n(\theta)}{T}\right)
	\end{equation}}
\end{enumerate}
These functions are sketched in Figure \ref{functionPicture}. They have the desired properties that $w=0$ when $n(\theta)=0$ (when no evaluations have been conducted that match the tuple), and $w \to 1$ as $n(\theta) \to \infty$. They differ in the rate at which this decay happens, which in all cases must be parameterised by some $T$. The Linear decay is most draconian, and will ignore any information from lower level tuples once $n \ge T$, while under Inverse-root decay lower-level tuples have a residual weight of 0.71 after $T$ evaluations. 
For all experiments in this study we set $T=15$. This is somewhat arbitrary, but scaled to be about 5\% of the total iterations in the smallest experiments with a budget of about 300 NTBEA iterations.

\begin{table*}[t]
	\centering
	\begin{tabular}{l|c|c|c|c}
		\hline
		Parameter &Planet Wars I & Asteroids I & Planet Wars II & Asteroids II \\
		\hline
		Sequence Length & 5, 10, \textbf{15}, 20, 25, 30 & 5, 10, 15, 20, 50, \textbf{100}, 150  &
		7, 10, 13, 16, 20, 25, 30  & 50, 75, 100, 125, 150, 200 \\
		Mutated Points & 0, 1, 2, \textbf{3} & 0, 1, \textbf{2, 3}  &  1, 2, 3, 5, 10, 15, 20  &  1, 2, 3, 5, 10, 20, 30, 50 \\
		Resample & \textbf{1}, 2, 3 & 1, 2, 3  & \textbf{1}, 2, 3 & 1, 2, 3 \\
		Flip One Value & false, \textbf{true} & false, \textbf{true} & false, true  & false, true  \\
		Use Shift Buffer & false, \textbf{true} & false, \textbf{true} & false, true  & false, true  \\
		Mutation Transducer & false & false &false, true & false, true\\
		Repeat Prob. & - & - & 0.2, 0.4, 0.6, 0.8 & 0.2, 0.4, 0.6, 0.8 \\
		Discount Factor & 1.0 & 1.0 & 1.0, 0.999, 0.99, 0.95, 0.9 & 1.0, 0.999, 0.99, 0.95, 0.9 \\
		\hline
		Parameter Space size & 228 & 336 & 23,520 & 23,040 \\
		\hline
	\end{tabular}
	\caption{Parameter space for RHEA in Planet Wars and Asteroids game experiments. The first two columns for the I experiments are as in  \cite{Lucas_Liu_Bravi_Gaina_Woodward_Volz_Perez-Liebana_2019}. The optimal values found for the games in that paper and in \cite{Lucas_Liu_Perez-Liebana_2018} are in bold.}
	\label{RHEASimple}
\end{table*}
\section{Method} \label{Method}
We apply each of the decay functions (\ref{linear}), (\ref{inverse}), (\ref{inverseRoot}), \ref{exponential}) to a number of different optimisation problems to determine whether our hypothesis holds and the modified model does converge faster and more robustly than vanilla NTBEA. By using a number of different problems we seek to test that any improvement generalises, and is not specific to one domain. A secondary goal is exploratory, to see if the four different weighting functions have varying patterns of performance.
\subsection{Benchmark functions}
We test on four benchmark functions from the global optimisation literature \cite{dixon1978global, Jones_Schonlau_Welch_1998}. These are interesting non-convex functions for which we can calculate the true value, and hence judge the performance for the NTBEA variants. Some amendments are needed to the original functions:
\begin{enumerate}
	\item {These are all deterministic functions with no noise. To convert them to a stochastic win/lose setting appropriate for a game benchmark we convert the function value to a probability $p$ of a +1 score (a `win'), and a 1-$p$ probability of a -1 score (a `loss'). }
	\item{They are continuous functions in all dimensions. We discretise by taking values at equally spaced intervals for each dimension.}
	\item{Global optimisation seeks to minimise a function. To maximise we multiply by -1.}
\end{enumerate}
 We outline the four functions below. A complete description is in \cite{dixon1978global}.
\begin{itemize}
	\item{\emph{Hartmann-3}. A three-dimensional function with four local optima. Two of these optima are close in value, with one slightly higher. In the original problem the output range is [0.0, 3.59], so we divide by 4.0 to get a $p$ value between 0 and 1. We split all three dimensions into ten equally spaced discrete values, for a total parameter-space size of 1000 with a true value $p \in [0, 0.897]$. } 
	\item{\emph{Hartmann-6}. A six-dimensional function with a similar four optima to \emph{Hartmann-3}. We apply the same modifications as with \emph{Hartmann-3}, and discretize each dimension into five equally spaced values, for a parameter-space of size 15,625 with $p \in [0, 0.737]$.}
	\item{\emph{Branin}. A two-dimensional function with three global maxima at 0.4. We split each dimension into 20 equally spaced intervals to get a parameter-space of 400. We add 10 to the result, divide by 12 with a floor at 0 to get to a valid range for $p \in [0, 0.795]$. In this case only 14.8\% of the 400 points are non-zero.}
	\item{\emph{Goldstein-Price}. A two-dimensional function with one global maximum, and several local ones. We split each dimension into 20 equally spaced intervals to get a parameter-space of 400. We add 400 to the result, divide by 500 with a floor at 0 to get to a valid range for $p \in [0, 0.794]$. 13.3\% of the 400 points are non-zero.}
\end{itemize}
In all cases we try each weighting function, plus vanilla NTBEA on each benchmark function with 300, 1000 and 3000 iterations. For each setting we run NTBEA 1000 times, and record the estimated $p$ value (by NTBEA) of the finally selected $\theta$ and the actual $p$ value.
In NTBEA we use $k=1$ for the exploration constant in (\ref{UCBEpsilon}).

\subsection{Game Parameters}
Lucas et al. 2019 \cite{Lucas_Liu_Bravi_Gaina_Woodward_Volz_Perez-Liebana_2019} compare NTBEA against several other popular optimisation algorithms in two games; Planet Wars and Asteroids. They optimise a Rolling Horizon Evolutionary Algorithm (RHEA) to find the best setting to win the 2-player Planet Wars (+1 for a win, and -1 for a loss), and also to obtain the highest score in 2000 game-ticks in the 1-player Asteroids. 
For comparable results we use exactly the same games and settings. In Planet Wars we use $k=1$ for the exploration constant in (\ref{UCBEpsilon}), and $k=5000$ for Asteroids.

In Planet Wars each player has a number of planets which generate ships at a constant rate. Players send ships from a planet to invade another, and to win the game they must conquer all planets. In Asteroids the player controls a ship which can rotate and shoot to destroy surrounding asteroids. Points are gained for shooting asteroids, and if one collides with the player then a life is lost; after three lost lives the game ends.
The details of the gameplay are not central to this study, and more details can be found in \cite{Lucas_Liu_Perez-Liebana_2018, Lucas_Liu_Bravi_Gaina_Woodward_Volz_Perez-Liebana_2019}.

RHEA is optimised over five parameters in \cite{Lucas_Liu_Bravi_Gaina_Woodward_Volz_Perez-Liebana_2019}, which are listed in Table \ref{RHEASimple}. Each optimisation algorithm was permitted 288 evaluations in Planet Wars, and 336 in Asteroids. This allowed Grid Search to run one game for each parameter setting. 
We repeat these experiments up to 100 times for each game and each weighting function. We record the parameter setting that is chosen each time.
To get a good estimate of the actual value of the 288 and 336 possible settings it is feasible to run 1000 games for each setting of Planet Wars and 500 for Asteroids, although this takes 6 days to run for Asteroids, illustrating the value of a rapid optimiser.

These small parameter spaces of 228 and 336 have the advantage of permitting a good estimate of the `best' setting to be found by brute force computation, but they are not representative of larger spaces in real problems. For example when optimising RHEA for a Game of Life variant \cite{lucas2019local} use NTBEA with 100 evaluations in a space of size 28,800. 
As a final experimental set we add further parameters to RHEA (discount factor, mutation transducer and repeat probability) from \cite{lucas2019local}, and extend the other parameters to give a larger overall space as detailed in Table \ref{RHEASimple} in the `II' columns. These extensions were fixed after seeing the results of the first set of experiments (the `I' columns) to focus on areas with higher performance. 
For Planet Wars we increased the concentration of Sequence Length options around the optimal 10-15 range, and in Asteroids we did the same around the optimal 100 value. We also increased the upper range of Mutated Points significantly, especially for Asteroids where the optimal value of 3 was the highest possible.

For these larger parameter spaces we used a budget of about 20,000 total iterations to try different overall approaches:
\begin{itemize}
	\item {10 runs of 2,000 iterations each}
	\item {3 runs of 7,000 iterations each}
	\item {2 runs of 10,000 iterations each}
	\item {1 run of 20,000 iterations}
\end{itemize}
Given the size of the parameters spaces it was not feasible to estimate an accurate value for all parameter settings. Instead we do this (by running 1000 or 500 games for Planet Wars and Asteroids respectively) for just the settings suggested by any of these runs. 
The purpose of these experiments is to understand how best to spend an available budget of iterations. Should we use them in a single NTBEA run, or spread them out and then pick the best of the suggestions. This is motivated by an observation from Deep Reinforcement Learning research, in which the random seed can have a major effect on the outcome of the algorithm, and results are often reported using `best of N' runs \cite{Henderson_Islam_Bachman_Pineau_Precup_Meger_2018}.

\section{Results} \label{Results}
\subsection{Benchmark functions}
\begin{figure*}[!t] 
	\centering
	\includegraphics[width=7in]{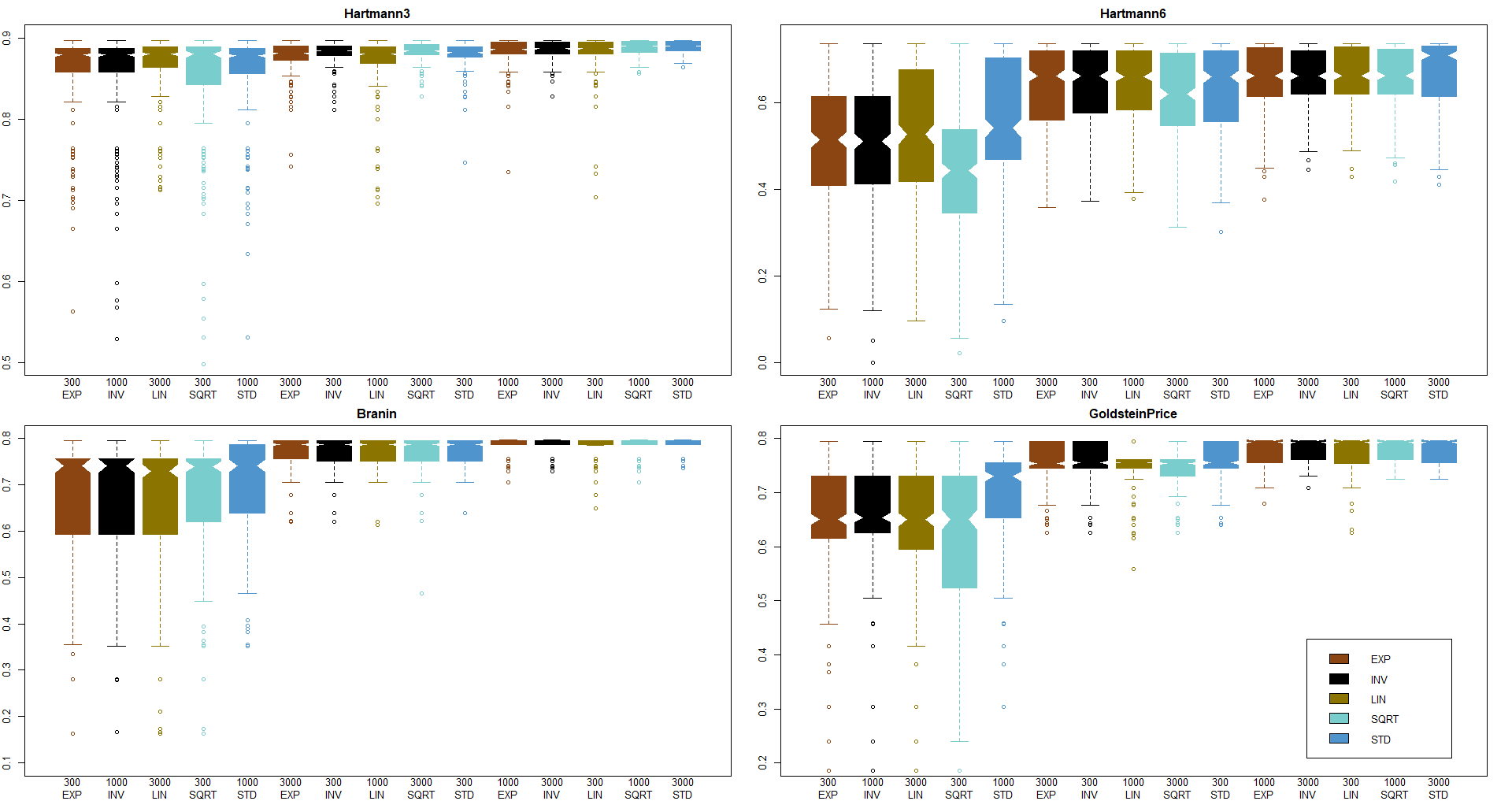}
	\caption{Boxplots for the true Score of settings recommended by NTBEA after 300, 1000 and 3000 iterations in each of the four benchmark functions.}
	\label{fig:FunctionPlot}
\end{figure*}

Table \ref{FunctionResults} in the Supplementary Material tabulates the numeric means and confidence intervals for the NTBEA experiments on the four benchmark functions with added noise. 
Figure~\ref{fig:FunctionPlot} displays boxplots of the true value of the NTBEA recommended parameters for each benchmark function and weighting function (1000 NTBEA runs for each, at 300, 1000 and 300 iterations).
\begin{itemize}
	\item {Hartmann-3. The appears to be the easiest of the four functions for NTBEA to optimise, with 300 iterations getting a mean value of 0.862 of a maximum of 0.897 for both Vanilla NTBEA (STD), and the Linear and Inverse-root weighting functions. With 3000 iterations all of the variants obtain a mean score of between 0.88 and 0.89; in all cases 25\% to 35\% of all runs recommend one of the three top parameter settings with actual values between 0.895 and 0.897}
	\item{Hartmann-6. This is harder to optimise with a clear progression as iterations increase from 300 to 3000. Vanilla NTBEA is a clear winner at only 300 iterations, and the Inverse-root and Inverse weighting functions are joint top with the Vanilla version at 3000 iterations (in a parameter space of size 15,625). The Linear weighting function does very poorly in comparison. }
	\item{Branin. As with Hartmann-6, Vanilla NTBEA is a clear winner at 300 iterations, and is joint top with the Inverse-root and Inverse weighting functions at 3000 iterations. The parameter space is only 400.}
	\item{Goldstein-Price. The same pattern is repeated here. Vanilla NTBEA is best for a small number of iterations, and all except the Linear weighting function are equally good with 3000 iterations to explore a parameter space of size 400.}
\end{itemize}
The key finding is that here vanilla NTBEA (`STD' in Figure~\ref{fig:FunctionPlot}) is always the best or joint best for any combination of benchmark function and number of iterations, and is particularly effective for smaller numbers of iterations.

\subsection{Games}
\begin{table*}[]
	\centering
	\begin{tabular}{lrrlrrrrrrrr}
		\hline
		NTBEA & Runs & Iterations & Game & Mean  &  S Dev   & \multicolumn{2}{l}{95\% Interval} & Delta  & \multicolumn{2}{l}{95\% Interval} & Top6 \\
		\hline
		STD   & 100  & 288        & Planet Wars   & \textbf{0.655} & 0.079 & 0.640           & 0.671           & -0.185 & -0.203          & -0.167          & 60\%   \\
		LIN   & 100  & 288        & Planet Wars   & 0.615 & 0.111 & 0.593           & 0.636           & 0.187  & 0.164           & 0.211           & 44\%   \\
		INV   & 100  & 288        & Planet Wars   & \textbf{0.630} & 0.110 & 0.610           & 0.656           & 0.086  & 0.061           & 0.107           & 53\%   \\
		SQRT  & 100  & 288        & Planet Wars   & 0.633 & 0.097 & 0.616           & 0.653           & \textbf{-0.017} & -0.036          & 0.001           & 51\%   \\
		EXP   & 100  & 288        & Planet Wars   & \textbf{0.643} & 0.091 & 0.625           & 0.663           & 0.130  & 0.107           & 0.151           & 58\%  \\
		\hline
		STD  & 62 & 336 & Asteroids & \textbf{9596}      & 67          & 9580 & 9613 & -709 & -741 & -680 & 94\% \\
		LIN  & 66 & 336 & Asteroids & \textbf{9577}      & 87 & 9556 & 9598 & 129  & 104  & 155  & 88\% \\
		INV  & 68 & 336 & Asteroids & \textbf{9584}      & 77          & 9567 & 9604 & \textbf{-27}  & -52  & -3   & 87\% \\
		SQRT & 69 & 336 & Asteroids & 9563      & 118         & 9536 & 9591 & -248 & -274 & -222 & 81\% \\
		EXP  & 67 & 336 & Asteroids & 9570      & 82 & 9552 & 9590 & 104  & 80   & 125  & 87\% \\
		\hline
	\end{tabular}
\caption{Results for Weighted NTBEA variants with Planet Wars and Asteroids.  Mean is the \emph{estimated} value of the final recommended parameter setting from 1000 offline games, with a 95\% confidence interval. Delta is the average difference to the NTBEA-estimated value of this point in the N-Tuple model, with a 95\% confidence interval. All confidence intervals are calculated with a basic bootstrap. Bold entries indicate the best performing variants (within confidence bounds) for each game. LIN is the Linear weighting function; INV is Inverse, SQRT is the Inverse-root and EXP is the Exponential.  STD is vanilla NTBEA. Top6 is the percentage runs that recommended one of the Top 6 parameter settings as estimated from the 1000 games run for each.}
\label{smallGameResults}
\end{table*}

\begin{figure}[!t] 
	\centering
	\includegraphics[width=3in]{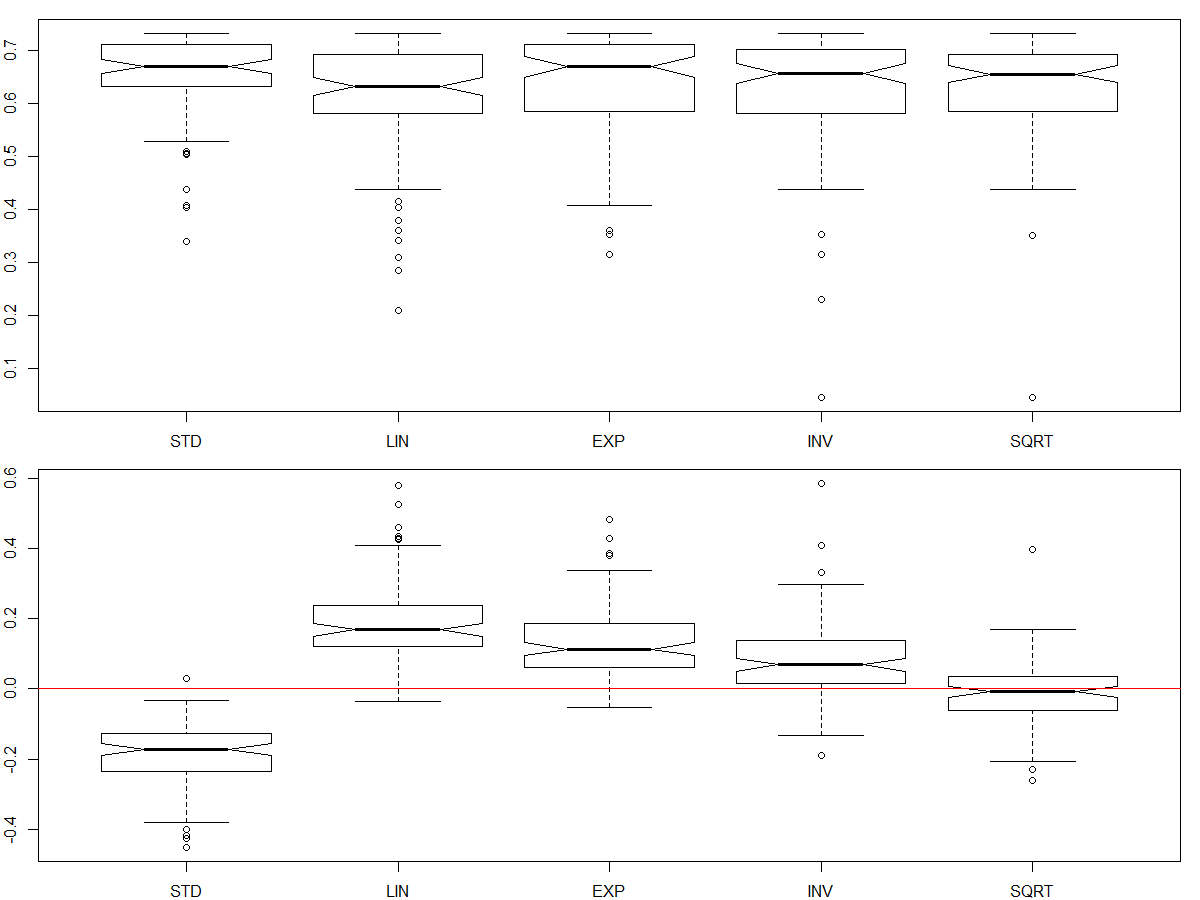}
	\caption{Boxplots for the estimated true Score of settings recommended by NTBEA after 288 iterations in Planet Wars (top), and the Delta of the NTBEA predicted value to this (bottom). The red horizontal line marks a Delta of 0.0, indicating perfect prediction by NTBEA.}
	\label{fig:PW}
\end{figure}

\begin{figure}[!t] 
	\centering
	\includegraphics[width=3in]{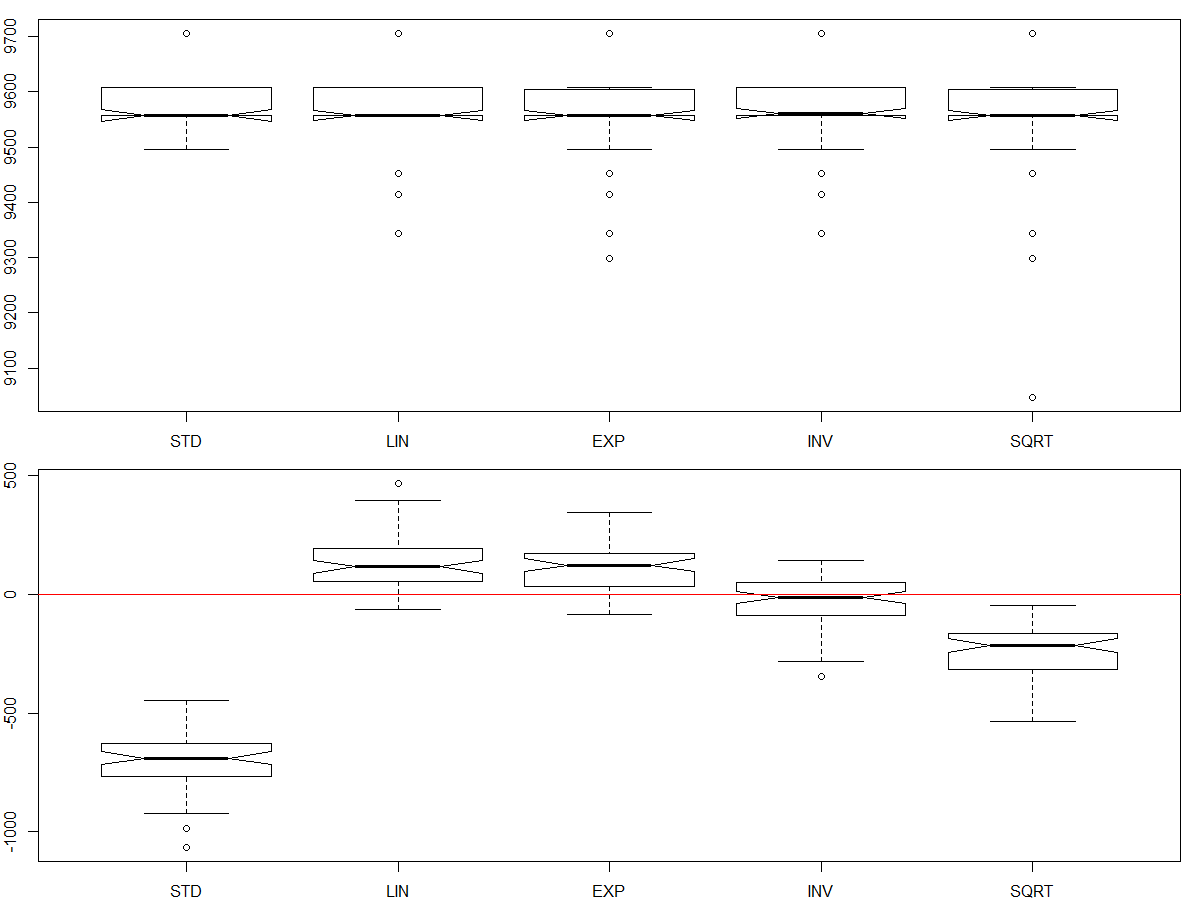}
	\caption{Boxplots settings recommended by NTBEA after 336 iterations in Asteroids. Key as in Figure~\ref{fig:PW}.}
	\label{fig:Ast}
\end{figure}

Table \ref{smallGameResults} shows the results for Planet Wars I and Asteroids I experiments, with 228 and 336 NTBEA iterations on similarly sized parameter spaces. Figures~\ref{fig:PW} and~\ref{fig:Ast} have box plots for the data. These are averaged over 100 runs for each setting for Planet Wars, and between 62 and 69 runs for Asteroids (the number that completed in an 84 hour window). For Planet Wars vanilla NTBEA gives both the best and most reliable (i.e. lowest standard deviation) results. The Exponential decay variant is the only one to have a performance within the 95\% confidence interval of vanilla NTBEA. 
The single highest parameter setting gives a score of 0.732, with 6 of the 288 settings having a score of 0.65 or higher averaged over 1000 games. Since we have run 1000 games for each of the 288 settings and \emph{then} picked the highest result, the 0.732 will be an over-estimate. Apart from the Linear weighted variant, all algorithms pick one of the top 6 settings between 50\% and 60\% of the time. 

For Asteroids the results are quite similar. Vanilla NTBEA gives the best result with the smallest standard deviation. One of the variants is within the 95\% confidence interval, but in this case it is the Inverse weighting function. In both games is is clear, as in the Benchmark Function results, that vanilla NTBEA gives the best recommended parameter setting despite giving a very poor estimate of the absolute value that the recommendation will provide when used.

The 95\% confidence intervals in Table \ref{smallGameResults} are calculated on the basis that the estimated values of each parameter setting are exact. This was true for the benchmark functions in Table \ref{FunctionResults}, but is not true here due to noise in these estimates from averaging across 1000 or 500 independent games. We do not have an estimate of this additional uncertainty.

Encouragingly, we obtain exactly the same the optimal parameter settings for both games as those found in the original work (highlighted in Table \ref{RHEASimple}) \cite{Lucas_Liu_Bravi_Gaina_Woodward_Volz_Perez-Liebana_2019, Lucas_Liu_Perez-Liebana_2018}. However, we get rather higher values for these in game play. For Planet Wars the original work finds that 288 iterations of NTBEA achieves a score of $0.51 \pm 0.01$, while we obtain $0.65$. In Asteroids the relevant values are $8,760 \pm 40$, against our $9600$. The reason for this discrepancy is not clear, but we do not believe it affects the key conclusions of this study.


\begin{table*}[]
	\centering
	\begin{tabular}{lllllllll}
		\hline
		Game        & NTBEA & Iterations & Runs & Best score     & Mean           & SD    & \multicolumn{2}{l}{95\% Bounds} \\
		\hline
Planet Wars & STD   & 1000       & 20   & 0.772      & \textbf{0.707} & 0.045 & 0.688  & 0.727 \\
Planet Wars & LIN   & 1000       & 20   & 0.752      & \textbf{0.679}          & 0.067 & 0.652  & 0.711 \\
Planet Wars & INV   & 1000       & 20   & 0.788      & \textbf{0.694}          & 0.070 & 0.665  & 0.728 \\
Planet Wars & SQRT  & 1000       & 20   & 0.762      & \textbf{0.712} & 0.035 & 0.697  & 0.728 \\
Planet Wars & EXP   & 1000       & 20   & 0.774      & \textbf{0.681}          & 0.061 & 0.656  & 0.708 \\
Planet Wars & STD   & 3000       & 7    & 0.762      & 0.709          &  &   & \\
Planet Wars & LIN   & 3000       & 7    & 0.762      & 0.718          & &   &  \\
Planet Wars & INV   & 3000       & 7    & 0.762      & 0.708          & &   &  \\
Planet Wars & SQRT  & 3000       & 7    & 0.760      & 0.714          & &   &  \\
Planet Wars & EXP   & 3000       & 7    & 0.774      & \textbf{0.735} &  &  &  \\
Planet Wars & STD   & 10000      & 2    & 0.756      & 0.717          &       &        &       \\
Planet Wars & LIN   & 10000      & 2    & 0.748      & 0.736          &       &        &       \\
Planet Wars & INV   & 10000      & 2    & 0.756      & 0.747          &       &        &       \\
Planet Wars & SQRT  & 10000      & 2    & 0.756      & 0.740          &       &        &       \\
Planet Wars & EXP   & 10000      & 2    & 0.770      & 0.748          &       &        &       \\
Planet Wars & STD   & 20000      & 1    & 0.708      &                &       &        &       \\
Planet Wars & LIN   & 20000      & 1    & 0.640      &                &       &        &       \\
Planet Wars & INV   & 20000      & 1    & 0.674      &                &       &        &       \\
Planet Wars & SQRT  & 20000      & 1    & 0.732      &                &       &        &       \\
Planet Wars & EXP   & 20000      & 1    & 0.632      &                &       &        &      \\
		\hline
Asteroids & STD  & 1000  & 20 & 9815          & \textbf{9701} & 63  & 9675 & 9728 \\
Asteroids & LIN  & 1000  & 20 & 9776		& 9655 & 89  & 9617 & 9694 \\
Asteroids & INV  & 1000  & 20 & 9803          & \textbf{9706} & 70  & 9690 & 9722 \\
Asteroids & SQRT & 1000  & 20 & 9811          & \textbf{9702} & 68  & 9676 & 9736 \\
Asteroids & EXP  & 1000  & 20 & 9819 & 9620 & 125 & 9569 & 9673 \\
Asteroids & STD  & 3000  & 7  & 9804          & 9707 &  &  &  \\
Asteroids & LIN  & 3000  & 7  & 9804          & 9764 &   &  &  \\
Asteroids & INV  & 3000  & 7  & 9835          & \textbf{9778} &   &  &  \\
Asteroids & SQRT & 3000  & 7  & 9817          & 9736 &   &  &  \\
Asteroids & EXP  & 3000  & 7  & 9818          & 9758 &   &  &  \\
Asteroids & STD  & 10000 & 2  & 9705 & 9705 &     &      &      \\
Asteroids & LIN  & 10000 & 2  & 9709          & 9612 &     &      &      \\
Asteroids & INV  & 10000 & 2  & 9804          & 9801 &     &      &      \\
Asteroids & SQRT & 10000 & 2  & 9817          & 9814 &     &      &      \\
Asteroids & EXP  & 10000 & 2  & 9779          & 9762 &     &      &      \\
Asteroids & STD  & 20000 & 1  & 9735          &      &     &      &      \\
Asteroids & LIN  & 20000 & 1  & 9783          &      &     &      &      \\
Asteroids & INV  & 20000 & 1  & 9783          &      &     &      &      \\
Asteroids & SQRT & 20000 & 1  & 9815          &      &     &      &      \\
Asteroids & EXP  & 20000 & 1  & 9815          &      &     &      &      \\
		\hline
	\end{tabular}
\caption{Results for Weighted NTBEA variants with Planet Wars and Asteroids over larger parameter spaces.  Mean is the \emph{estimated} value of the final recommended parameter setting from 1000/500 offline games for Planet Wars/Asteroids, with 95\% confidence intervals calculated with a basic bootstrap. Bold entries indicate the best performing variants (within confidence bounds) for each game. Best score is the best individual result for any of the runs for that line.}
\label{largeGameResults}
\end{table*}
Table \ref{largeGameResults} shows the results from the Planet Wars II and Asteroids II experiments with larger, more realistic parameter spaces to explore. 
There were 142 unique parameter settings recommended by the 150 NTBEA runs for the Planet Wars II experiments and an estimated value for each of these was calculated from averaging 1000 runs of the game.
The best estimated scores of the recommended parameter settings have increased to 0.77 compared to the best possible score of 0.73 for Planet Wars I, so the additional parameters enable RHEA to better play the game if we can efficiently explore the space. 

For Planet Wars vanilla NTBEA gives the best mean result at 1k iterations, and does not give significantly different results at more iterations (within 95\% error bounds).
The same caveat applies to these error bounds as in Table \ref{smallGameResults} as they do not include the additional uncertainty from the average over 1000 runs used to estimate the value of the final parameter settings.
 
The Inverse-root weighting functions matches vanilla NTBEA at 1k, and at 3k all variants at least match vanilla performance, with the Exponential weighting being the best. 
These results make clear that there is a high level of uncertainty in any individual NTBEA run. The best of the 20 vanilla runs at 1k gives a parameter setting that scores 0.772 over 1000 games, and the worst scores a mere 0.616. This remains true at 10k and 20k iterations, with three of the 20k runs recommending parameters that score less than 0.7.

Even with a large number of iterations any single NTBEA run may give a relatively poor result. Given a fixed budget of games to optimise a parameter Table \ref{largeGameResults} suggests that it is not a good idea to put the whole budget into a single NTBEA run. Far better to execute several NTBEA runs with a small number of iterations, and then use the remaining game budget to estimate the true value of each of these and pick the best. 

This is reinforced when we look at the Asteroids results in Table \ref{largeGameResults}. Vanilla NTBEA does joint best with 1k iterations, and the mean score does not increase significantly for higher numbers of iterations. At higher iterations all variants except the Linear function are at least as good, but not necessarily reliably better. In the Asteroids case there is an effective maximum score of 10000 when we use 2000 game ticks as here, so with all the mean and best results in the 9700 to 9800 range the optimisation does not have much room to work, especially when we add noise.

\section{Discussion}

In all four of the benchmark functions, and in both games across small and large parameter spaces vanilla NTBEA is at least as good as  the weighting variants tried for small numbers of iterations; and usually better with lower variance in results. As the number of iterations increases this effect shrinks, and for some cases one of the weighting variants can be significantly better. For example Inverse-root with 1000 iterations on the Hartmann-6 function, or the Exponential function with 3000 iterations in Asteroids II. However, this is cherry-picking.
Furthermore the weighting variants introduce complexity with a new hyper-parameter $T$ to be specified.

When we optimise an expensive function such as game performance over a parameter space we are deliberately trying to use a small number of iterations. Vanilla NTBEA works best in this situation, and we conclusively reject the hypothesis that improving the N-Tuple model with these weighting functions improves either reliability or performance. 

We do not reject the hypothesis that the variants provide a better estimate of the true value of a parameter setting. Across all benchmark functions and game environments vanilla NTBEA provides very poor estimates of the actual value, under-estimating by a very large margin because it is averaging over all possible Tuple matches. The Inverse and Inverse-root weighting functions consistently do a much better job of estimating the value of their recommendation. However, this is not as important when our key objective is to get a good recommendation; we can always go on to get a good estimate of its value later.

Linear weighting is clearly worse than the other options that do not exclude all contributions from less-specific Tuples with more information after only $T$ iterations. The appears to be because once it has $T$ evaluations of a specific setting it ignores all other data, and uses the average of those evaluations. With a larger number of iterations what often happens is that sequential iterations focus on the current best estimate until the mean falls sufficiently and the focus shifts to another setting. With noisy function evaluations this often leads to a recommendation with a smaller number of trials (but more than $T$), that happens to currently have a high estimate. Hence the recommendation is optimistic because it picks the best (stochastic) estimate across all options with more than $T$ evaluations, and we can see this reflected in the general over-estimate of the value of its recommendation (a version of the `winner's curse'). This effect is less evident for the other weighting functions, as they never let the weighting of other Tuples fall to zero.

\section{Conclusion and Future Work} \label{Conclusions}
We hypothesised that adding a recursive weighting function to apply to Tuples in NTBEA would improve performance in parameter optimisation in terms of quality and reliability of a recommended (optimised) parameter setting and in providing a more accurate estimate of the value of this. We tried four different weighting functions with different decay characteristics (linear, inverse, inverse-root and exponential) across four benchmark functions from the function optimisation literature, and two games with two distinct sizes of parameter space.

Across all ten experiments we found no evidence that the proposed weighting functions improved NTBEA except in the least important one of providing a better estimate of the true value of the parameter setting recommended by the optimising process. On the contrary, we found strong evidence that vanilla NTBEA is better able than the weighting function variants to reliably find a higher quality recommendation. This is especially true for the smaller number of iterations that would tend to be used in real world applications.

Finally we investigated how best to use a fixed budget of NTBEA iterations in the Planet Wars and Asteroids games. These showed than any individual NTBEA run may give a poor recommendation, and it is better to run several NTBEA runs with a smaller number of iterations, and then use the remaining budget to estimate more accurately the value of these, and then pick the best.

We have not explored different values of $T$, the hyper-parameter introduced to determine how the weighting function is used, and it is possible that other values may perform better.
There are other more adventurous options to improve the N-Tuple model, such as regression across the tuples to determine which ones are important. The updated model in this paper still assumes that each Tuple at a given level is equally important. If we have no data for the full $d$-Tuple then we average across all matching 2-Tuples, when in practise some of these may be more important than others. One approach to try would be to construct a regression model across the tuples to up-weight the ones that better predict the observed results.
We have also not changed the exploration model, which averages across all matching tuples as in vanilla NTBEA. It could be worthwhile to experiment with different noise models, for example using a square root instead of a log function in Equation (\ref{UCB}), which has been found useful in other areas where exploration is more important than exploitation \cite{Tolpin_Shimony_2012}.

\section{Acknowledgments}
This work was funded by the EPSRC CDT in Intelligent Games and Game Intelligence (IGGI) EP/S022325/1.

%
%
%
%
%
\balance

\bibliographystyle{AISB2008}
\bibliography{sample}

\beginsupplement
\begin{table*}[]
	\centering
	\begin{tabular}{lrrl|cccc|ccc}
		\hline
		NTBEA & Runs & Iteration & Function       & Mean           & SD    & \multicolumn{2}{c|}{95\%  Bounds} & Delta  & \multicolumn{2}{c}{95\% Bounds} \\
		\hline
		STD   & 1000 & 300       & Hartmann-3      & \textbf{0.859} & 0.051 & 0.856                & 0.862               & -0.056          & -0.061               & -0.052              \\
		LIN   & 1000 & 300       & Hartmann-3      & \textbf{0.862} & 0.104 & 0.855                & 0.868               & 0.080           & 0.074                & 0.087               \\
		INV   & 1000 & 300       & Hartmann-3      & 0.855          & 0.057 & 0.852                & 0.859               & -0.075          & -0.084               & -0.066              \\
		SQRT  & 1000 & 300       & Hartmann-3      & 0.848          & 0.066 & 0.844                & 0.853               & -0.235          & -0.241               & -0.230              \\
		EXP   & 1000 & 300       & Hartmann-3      & \textbf{0.862}          & 0.044 & 0.859                & 0.865               & \textbf{-0.003} & -0.012               & 0.006               \\
		\hline
		STD   & 1000 & 1000      & Hartmann-3      & 0.881          & 0.016 & 0.880                & 0.882               & -0.075          & -0.078               & -0.072              \\
		LIN   & 1000 & 1000      & Hartmann-3      & 0.872          & 0.032 & 0.870                & 0.874               & 0.087           & 0.084                & 0.091               \\
		INV   & 1000 & 1000      & Hartmann-3      & 0.881          & 0.016 & 0.879                & 0.882               & 0.046           & 0.044                & 0.049               \\
		SQRT  & 1000 & 1000      & Hartmann-3      & \textbf{0.883} & 0.013 & 0.882                & 0.883               & \textbf{0.007}  & 0.004                & 0.010               \\
		EXP   & 1000 & 1000      & Hartmann-3      & 0.879          & 0.018 & 0.878                & 0.880               & 0.067           & 0.065                & 0.070               \\
		\hline
		STD   & 1000 & 3000      & Hartmann-3      & \textbf{0.888} & 0.009 & 0.887                & 0.888               & \textbf{-0.021} & -0.022               & -0.020              \\
		LIN   & 1000 & 3000      & Hartmann-3      & 0.882          & 0.022 & 0.881                & 0.884               & 0.042           & 0.040                & 0.044               \\
		INV   & 1000 & 3000      & Hartmann-3      & 0.886          & 0.010 & 0.885                & 0.886               & 0.031           & 0.029                & 0.032               \\
		SQRT  & 1000 & 3000      & Hartmann-3      & \textbf{0.888} & 0.009 & 0.887                & 0.888               & \textbf{0.020}  & 0.018                & 0.021               \\
		EXP   & 1000 & 3000      & Hartmann-3      & 0.885          & 0.012 & 0.884                & 0.886               & 0.037           & 0.035                & 0.039               \\
		\hline
		STD   & 1000 & 300       & Hartmann-6      & \textbf{0.551} & 0.142 & 0.542                & 0.560               & -0.127          & -0.135               & -0.118              \\
		LIN   & 1000 & 300       & Hartmann-6      & 0.535          & 0.142 & 0.526                & 0.544               & 0.119           & 0.107                & 0.130               \\
		INV   & 1000 & 300       & Hartmann-6      & 0.516          & 0.149 & 0.506                & 0.525               & \textbf{0.026}  & 0.015                & 0.038               \\
		SQRT  & 1000 & 300       & Hartmann-6      & 0.461          & 0.155 & 0.451                & 0.471               & -0.120          & -0.133               & -0.108              \\
		EXP   & 1000 & 300       & Hartmann-6      & 0.526          & 0.149 & 0.516                & 0.535               & 0.065           & 0.053                & 0.077               \\
		\hline
		STD   & 1000 & 1000      & Hartmann-6      & 0.633          & 0.095 & 0.627                & 0.639               & -0.178          & -0.188               & -0.168              \\
		LIN   & 1000 & 1000      & Hartmann-6      & 0.633          & 0.092 & 0.628                & 0.639               & 0.108           & 0.100                & 0.115               \\
		INV   & 1000 & 1000      & Hartmann-6      & \textbf{0.639} & 0.085 & 0.634                & 0.645               & 0.046           & 0.040                & 0.052               \\
		SQRT  & 1000 & 1000      & Hartmann-6      & 0.610          & 0.101 & 0.604                & 0.617               & \textbf{-0.031} & -0.041               & -0.020              \\
		EXP   & 1000 & 1000      & Hartmann-6      & 0.633          & 0.092 & 0.627                & 0.639               & 0.082           & 0.076                & 0.088               \\
		\hline
		STD   & 1000 & 3000      & Hartmann-6      & \textbf{0.666} & 0.085 & 0.661                & 0.672               & -0.202          & -0.220               & -0.184              \\
		LIN   & 1000 & 3000      & Hartmann-6      & 0.635          & 0.104 & 0.628                & 0.641               & 0.100           & 0.092                & 0.107               \\
		INV   & 1000 & 3000      & Hartmann-6      & \textbf{0.666} & 0.066 & 0.661                & 0.670               & 0.031           & 0.027                & 0.036               \\
		SQRT  & 1000 & 3000      & Hartmann-6      & \textbf{0.668} & 0.057 & 0.664                & 0.671               & \textbf{-0.006} & -0.012               & 0.000               \\
		EXP   & 1000 & 3000      & Hartmann-6      & 0.658          & 0.079 & 0.653                & 0.663               & 0.058           & 0.054                & 0.063               \\
		\hline
		STD   & 1000 & 300       & Branin         & \textbf{0.705} & 0.098 & 0.699                & 0.712               & \textbf{-0.020} & -0.027               & -0.013              \\
		LIN   & 1000 & 300       & Branin         & 0.676          & 0.142 & 0.667                & 0.685               & -0.389          & -0.398               & -0.380              \\
		INV   & 1000 & 300       & Branin         & 0.670          & 0.136 & 0.661                & 0.679               & -0.389          & -0.398               & -0.380              \\
		SQRT  & 1000 & 300       & Branin         & 0.460          & 1.327 & 0.376                & 0.544               & -0.204          & -0.288               & -0.120              \\
		EXP   & 1000 & 300       & Branin         & 0.678          & 0.126 & 0.670                & 0.686               & -0.397          & -0.405               & -0.388              \\
		\hline
		STD   & 1000 & 1000      & Branin         & \textbf{0.773} & 0.027 & 0.771                & 0.775               & -0.031          & -0.035               & -0.028              \\
		LIN   & 1000 & 1000      & Branin         & 0.768          & 0.035 & 0.766                & 0.771               & 0.055           & 0.050                & 0.060               \\
		INV   & 1000 & 1000      & Branin         & \textbf{0.772} & 0.026 & 0.771                & 0.774               & \textbf{0.015}  & 0.011                & 0.019               \\
		SQRT  & 1000 & 1000      & Branin         & \textbf{0.773} & 0.028 & 0.771                & 0.775               & -0.047          & -0.052               & -0.042              \\
		EXP   & 1000 & 1000      & Branin         & \textbf{0.773} & 0.029 & 0.771                & 0.775               & 0.041           & 0.037                & 0.044               \\
		\hline
		STD   & 1000 & 3000      & Branin         & \textbf{0.789} & 0.012 & 0.788                & 0.790               & -0.020          & -0.021               & -0.018              \\
		LIN   & 1000 & 3000      & Branin         & 0.781          & 0.024 & 0.780                & 0.783               & 0.024           & 0.021                & 0.026               \\
		INV   & 1000 & 3000      & Branin         & \textbf{0.789} & 0.013 & 0.788                & 0.789               & 0.011           & 0.009                & 0.012               \\
		SQRT  & 1000 & 3000      & Branin         & \textbf{0.788} & 0.015 & 0.787                & 0.789               & \textbf{-0.002} & -0.004               & 0.000               \\
		EXP   & 1000 & 3000      & Branin         & 0.784          & 0.020 & 0.783                & 0.785               & 0.019           & 0.017                & 0.022               \\
		\hline
		STD   & 1000 & 300       & GoldsteinPrice & \textbf{0.700} & 0.076 & 0.695                & 0.705               & \textbf{-0.005} & -0.011               & 0.001               \\
		LIN   & 1000 & 300       & GoldsteinPrice & 0.621          & 0.145 & 0.612                & 0.630               & -0.318          & -0.328               & -0.308              \\
		INV   & 1000 & 300       & GoldsteinPrice & 0.622          & 0.142 & 0.613                & 0.631               & -0.323          & -0.333               & -0.313              \\
		SQRT  & 1000 & 300       & GoldsteinPrice & 0.622          & 0.142 & 0.613                & 0.631               & -0.350          & -0.360               & -0.340              \\
		EXP   & 1000 & 300       & GoldsteinPrice & 0.614          & 0.142 & 0.605                & 0.623               & -0.313          & -0.323               & -0.303              \\
		\hline
		STD   & 1000 & 1000      & GoldsteinPrice & \textbf{0.759} & 0.029 & 0.757                & 0.761               & \textbf{-0.017} & -0.021               & -0.014              \\
		LIN   & 1000 & 1000      & GoldsteinPrice & 0.755          & 0.035 & 0.753                & 0.758               & 0.071           & 0.066                & 0.075               \\
		INV   & 1000 & 1000      & GoldsteinPrice & 0.756          & 0.029 & 0.754                & 0.758               & 0.024           & 0.020                & 0.028               \\
		SQRT  & 1000 & 1000      & GoldsteinPrice & 0.750          & 0.030 & 0.748                & 0.752               & -0.042          & -0.046               & -0.037              \\
		EXP   & 1000 & 1000      & GoldsteinPrice & 0.756          & 0.032 & 0.754                & 0.758               & 0.057           & 0.053                & 0.060               \\
		\hline
		STD   & 1000 & 3000      & GoldsteinPrice & \textbf{0.779} & 0.020 & 0.778                & 0.781               & -0.016          & -0.018               & -0.014              \\
		LIN   & 1000 & 3000      & GoldsteinPrice & 0.775          & 0.026 & 0.774                & 0.777               & 0.024           & 0.021                & 0.026               \\
		INV   & 1000 & 3000      & GoldsteinPrice & \textbf{0.780} & 0.020 & 0.778                & 0.781               & 0.014           & 0.012                & 0.016               \\
		SQRT  & 1000 & 3000      & GoldsteinPrice & \textbf{0.778} & 0.021 & 0.777                & 0.779               & \textbf{-0.001} & -0.003               & 0.001               \\
		EXP   & 1000 & 3000      & GoldsteinPrice & \textbf{0.778} & 0.022 & 0.777                & 0.780               & 0.020           & 0.018                & 0.022           \\   
		\hline
	\end{tabular}
	\caption{Results for Weighted NTBEA with benchmark functions. Mean is the \emph{actual} value of the optimised parameters. Delta is the difference between actual and N-Tuple estimated value of these parameters. Bold entries are the best performing variants (within confidence bounds) for each combination of function and number of iterations. LIN is the Linear weighting function; INV is Inverse, SQRT is the Inverse-root and EXP is the Exponential.  STD is vanilla NTBEA.}
	\label{FunctionResults}
\end{table*}
\end{document}